%% file: main_arxiv.tex
\documentclass[letterpaper, 10 pt, conference]{ieeeconf}  

\IEEEoverridecommandlockouts                              

\usepackage{adjustbox}
\usepackage[font=small,belowskip=-1.5pt]{caption} 
\usepackage{tikz}
\usepackage[colorlinks=true]{hyperref}

\usepackage{amssymb}
\usepackage{mathtools} 
\usepackage{bbm}

\usepackage{pifont}
\usepackage{alphalph}
\usepackage{amsmath}
\usepackage{booktabs}
\usepackage{multirow}
\usepackage{interval}
\usepackage{caption}
\usepackage{subcaption}

\definecolor{azure}{rgb}{0.0, 0.5, 1.0}

\newcommand{\ours}{VINet}
\newcommand{\rev}[1]{#1}

\begin{document}

\title{\LARGE \bf

VINet: Visual and Inertial-based Terrain Classification and Adaptive Navigation over Unknown Terrain

}

\author{
Tianrui Guan$^{1*}$, %
Ruitao Song$^{2\dag}$,
Zhixian Ye$^{2}$,
and Liangjun Zhang$^{2}$\\
\thanks{* Work done during an internship at Baidu RAL, USA.}
\thanks{$^{1}$Author is with the Department of Computer Science, University of Maryland, College Park, MD 20742, USA. Email: {\tt\footnotesize rayguan@umd.edu}}%
\thanks{$^{2}$Authors are with the Robotics and Autonomous Driving Laboratory, Baidu USA, Sunnyvale, CA 94089, USA. Emails: {\tt\footnotesize \{ruitaosong, zhixianye, liangjunzhang\}@baidu.com}}%
\thanks{\dag\ Corresponding author
}
}


\maketitle
\thispagestyle{empty}
\pagestyle{empty}

\begin{abstract}

We present a visual and inertial-based terrain classification network (\ours{}) for robotic navigation over different traversable surfaces. We use a novel navigation-based labeling scheme for terrain classification and generalization on unknown surfaces. Our proposed perception method and adaptive \rev{scheduling} control framework can make predictions according to terrain navigation properties and lead to better performance on both terrain classification and navigation control on known and unknown surfaces. Our \ours{} can achieve $98.37\%$ in terms of accuracy under supervised setting on known terrains and improve the accuracy by $8.51\%$ on unknown terrains compared to previous methods. We deploy \ours{} on a mobile tracked robot for trajectory following and navigation on different terrains, and we demonstrate an improvement of $10.3\%$ compared to a baseline controller in terms of RMSE.

\end{abstract}


\input{sections/intro}

\input{sections/related}
\input{sections/method}
\input{sections/result}

\begin{figure*}[t!]
    \centering
    \includegraphics[width=0.9\textwidth]{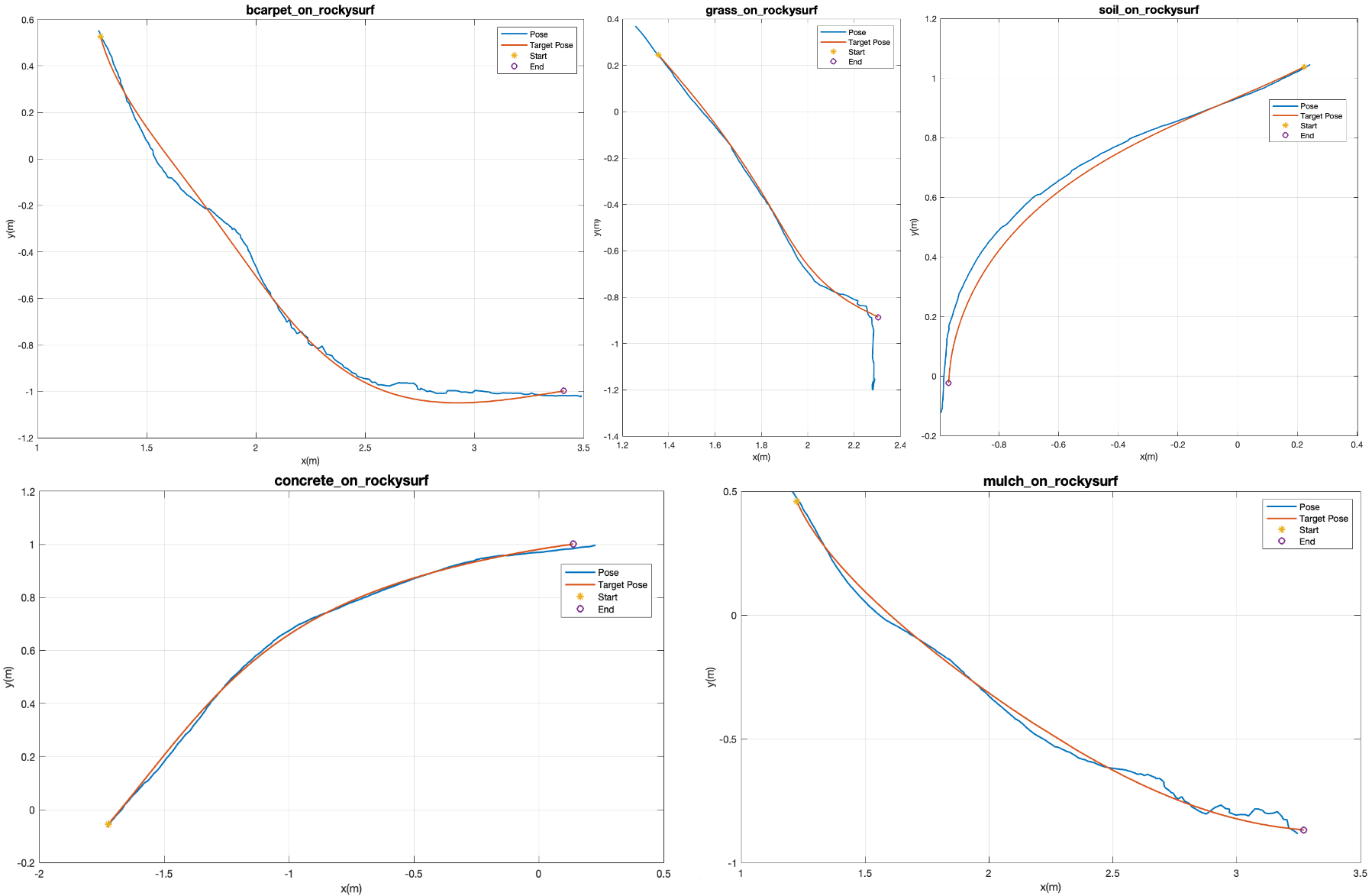}
    \caption{\textbf{Visualization of trajectories on rocky surface:} For each controller, we show one sample of the trajectories. We can see that the concrete controller can follow the target trajectory the best.}
    \label{fig:rocky}
\end{figure*}

\input{sections/conclusion}


\section*{APPENDIX}

\input{appendix/terrain_classes}


{\small
\bibliographystyle{IEEEtran}
\bibliography{citation}
}

\end{document}

%% file: sections/intro.tex
\section{Introduction}


The problems of perception and navigation have been extensively explored individually in the past. The perception tasks, including detection~\cite{swin}, segmentation~\cite{segformer, ocrnet}, classifications~\cite{coatnet, terrapn}, etc., can obtain very good overall accuracy running on challenging offline data sets~\cite{russakovsky2015imagenet, coco}, which are clean and well-annotated. For mobile robot navigation in the outdoor environment, efficient perception methods such as terrain traversability analysis~\cite{tns} have been proposed. Meanwhile, many control policies have been developed for mobile wheel or tracked robots to navigate on different terrains~\cite{mckinnon2018experience, dallas2021terrain}. 

Perception and navigation problems, however, are rarely considered as a whole. Many sophisticated perception models confine the task and problem settings for standard bench-marking and fair comparisons. To truly benefit some downstream tasks like navigation and control, we need to reconsider the definition and context of the perception algorithms that are the same as those in standard benchmarks. For example, categories defined in a traditional classification or semantic segmentation task might not be used directly in navigation settings, since we need a mapping between the original categories and traversability of the terrain. There are some works that focus on robotic perception for terrain understanding~\cite{ganav, offseg} and traversability analysis~\cite{geo1, tns}, but most of those definitions are hand-crafted and heuristic-based. Instead, metrics based on navigation could provide insight on how to define such mapping or labeling scheme.

Perception for outdoor navigation is challenging, especially in an environment that is unknown and constantly changing with uncertainty. The perception model needs to generalize to unseen terrain well since it's not possible to collect information and label all surfaces. In the context of navigation, unlike a traditional perception task, the perception algorithms should be trained in a way that maximizes the navigation performance and evaluated according to some navigation-related metrics, which is rarely explored in the past. The perception needs to be coupled with navigation and output the most useful result that can lead to stable and safe decision for the navigation and control modules.

\begin{figure}[t]
    \centering
    \includegraphics[width = \columnwidth]{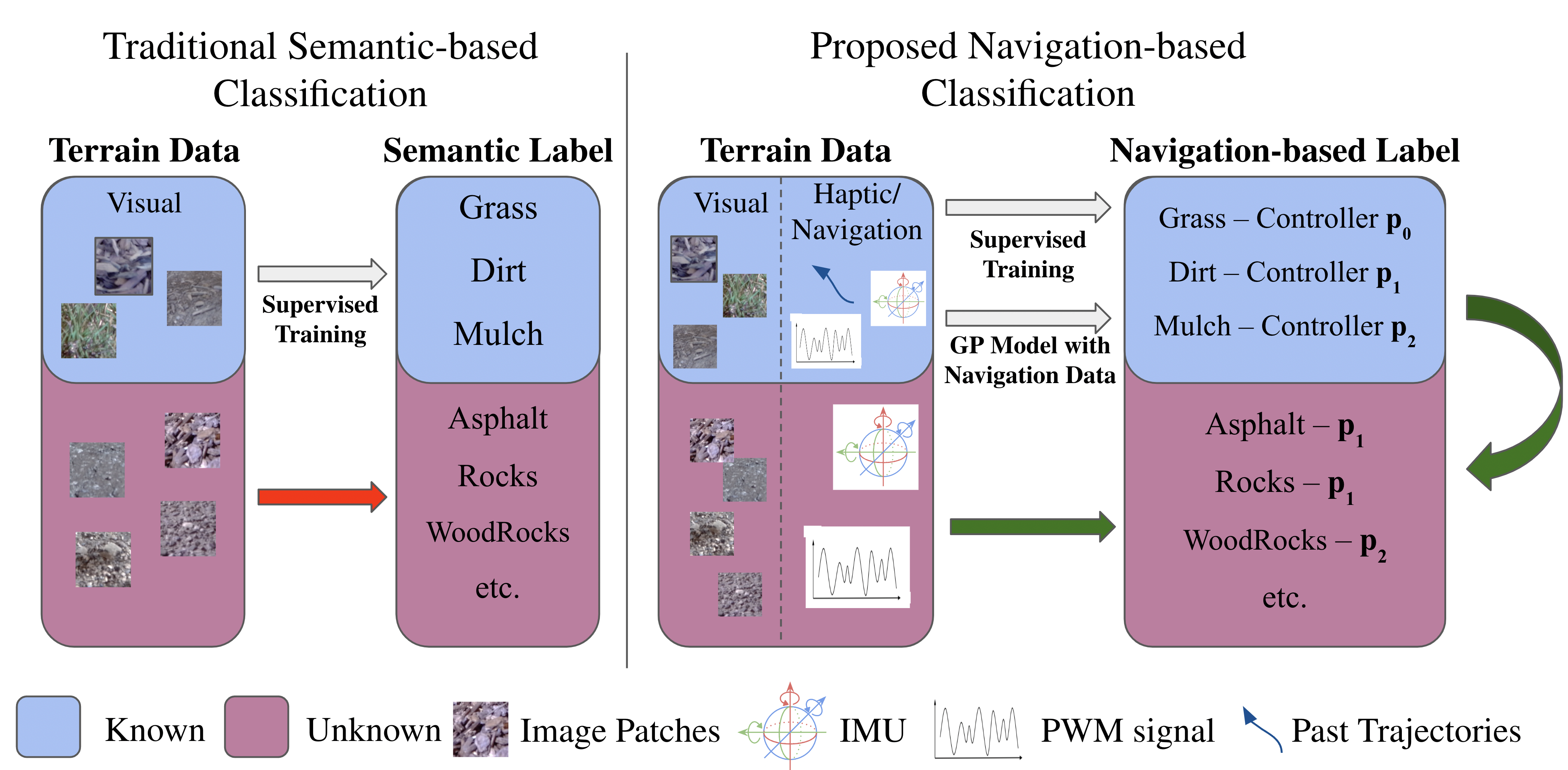}
    \caption{\textbf{Semantic labels v.s. Navigation-based labels:} The traditional classification (left) has semantic-based labels, which are unambiguous but not generalizable to unknown terrain because the prediction would be meaningless. Our proposed navigation-based classification (right) recasts the terrain labels to known classes based on navigation properties distinguished by different inertial feedback as part of the training data to train our proposed \ours{}, which can outperform other existing methods on both semantic-based and navigation-based labeling. }
    \vspace{-7mm}
    \label{fig:cover}
\end{figure}

In this paper, we propose \ours{}, a novel terrain classification network that combines both visual and inertial sensing through camera and Inertial Measurement Unit (IMU). Our method can work well on known terrains as well as completely unseen surfaces. There are two streams in the network: 1) an image stream that provides robust predictions on the training set and seen surfaces, and 2) an IMU temporal stream that can extract features of unseen terrains through inertial sensing and generalize to unknown classes. As shown in Figure~\ref{fig:cover}, we used a navigation-based perception metric to measure the performance of our network for unseen terrains, which maximizes the performance of navigation. In addition, we propose a novel adaptive \rev{scheduling} control framework and show that this pipeline with the proposed \ours{} leads to better navigation performance on both known and unknown terrain.

The key contributions of our work include:
\begin{enumerate}
    \item We propose a novel terrain classification network which fuses visual and inertial features of the terrain. Our method can achieve $98.37\%$ in terms of accuracy under the supervised setting and lead to an increase of $8.51\%$ under the unsupervised setting compared to previous methods.
    \item We propose a novel fusion scheme, which closes the latent embeddings between image and inertial data, and leverages those two modalities channel-wise. We also introduce an IMU denoising strategy to process noisy IMU data and further boost the classification performance.

    \item We deploy \ours{} on a tracked robot and propose a novel adaptive \rev{scheduling} control framework enabled by \ours{} with navigation-based label generalization. Our navigation performance outperforms a baseline single terrain controller by $10.3\%$ in terms of RMSE.

\end{enumerate}

%% file: sections/related.tex
\section{Related Work}


\subsection{Terrain Classification}


In \cite{8578163}, Xue et al. proposed an orderless-texture-based and spatial information-targeted neural network. 
\cite{Laible20123DLA} used both 3D LiDAR and camera to classify terrain by computing a 3D-scan feature that overcomes the issue in different lighting conditions.
Similarly, Nguyen \cite{nguyen} introduced a structure-based method as well as estimating the surface smoothness to classify different roughness surface objects.
\cite{8100217} proposed an angular imaging method to distinguish different material types.  


In addition, some methods try to solve the classification problem with inertial sensing.
The earlier method like \cite{ward2007terrain} utilizes a dynamic model to estimate the terrain profile and further label the terrain based on the spatial frequency part of the profile predicted. 
Likewise, \cite{ward2009speed} adopts the same frequency analysis to classify the terrain by filtering out the terrain impulses from the identified terrain profile.
\cite{articleHashmi} tries to reveal the terrain types using the inertial data collected from the human walk.
Recent work \cite{lomio2019surface} proposes several deep learning methods for classifying the time-series IMU data on top of the engineered features.  

\subsection{Navigation on different Terrain}

Indoor navigation has been well developed in the past decade \cite{khan2022recent, densecavoid, frozone}, but it still remains a challenge in the outdoor environment \cite{droeschel2017continuous, terp, adaptiveon}, especially on different terrains \cite{kumar2021rma, viikd}.
Kumar et al. \cite{kumar2021rma} investigate adaptive control strategies to enable a quadruped robot to walk on different material terrains, such as sand, mud, grass, and dirt. 
The wheeled robot or tracked robot \cite{lee2003double} is another type of platform that is used in terrain navigation.
\cite{vincent2012combined, zong2020angle} and \cite{ordonez2018modeling} cover the topics of unstructured environment modeling and controller design from the angle of the tracked robot. 
As mentioned in \cite{kumar2021rma}, the adaptive control policy enables the legged robot to quickly recover stability from falling, which poses a key factor towards the success of navigating in dynamics-changing environments of different terrain.
Such related learning-based adaptive controller methods have been discussed in
\cite{hewing2020learning, torrente2021data, 8793757}.

%% file: sections/method.tex
\section{Method}

\begin{figure*}[t]
    \centering
    \includegraphics[width=\textwidth]{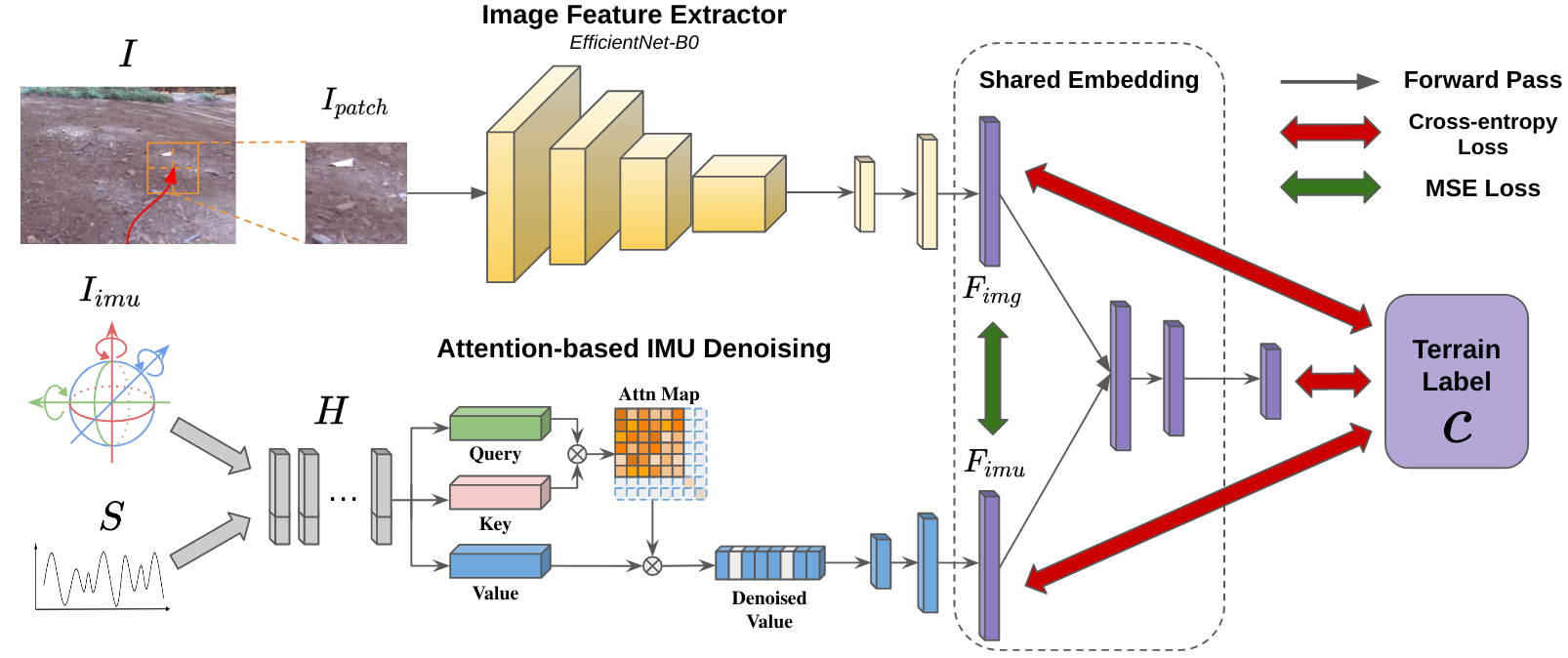}
    \caption{\textbf{Architecture of our proposed network \ours{}:} Our proposed network consists of two branches: an image branch and an IMU branch. We use EfficientNet-B0 as the image backbone, and for the IMU branch, we present a novel attention-based IMU denoising module. Each branch acts as a feature extractor and maps the visual and inertial data to a shared embedding space for the final prediction. 
    }
    \label{fig:network}
    \vspace{-15pt}
\end{figure*}

In this section, we present \ours{}, a novel terrain classification method for adaptive navigation and control. Our method combines visual and inertial information to make a classification prediction on different types of terrains and can classify known and unknown terrains categorized by navigation metrics. We later show that our network can directly benefit decision-making for control and navigation, even on terrains that the network has never seen before. 

The rest of the sections are structured as follows: We begin by defining our problem settings and each component of our network, including backbones, IMU denoising, classification head, and loss functions. We move on to our implementation of adaptive \rev{scheduling} control and the entire framework for navigation. Finally, we give some insights into the designs, including the navigation-based labeling and using IMU data as input for the classification task.

\subsection{Problem Definition and Terminology}
\label{def}
The inputs of the network consist of an image patch $I_{patch} \in \mathbb{R}^{3\times l \times l}$ taken along the intended trajectory of the mobile robot from the RGB camera, a sequence of 6 dimensional IMU data $I_{imu}(t_s) = [a_x, a_y, a_z, w_x, w_y, w_z](t_s) \in \mathbb{R}^{6}, t_s = \{t_0, ..., t_0 + D\}$ with three linear accelerations and three angular velocities for each axis, and a pulse-width modulation (PWM) signal $S(t_s) \in 
[0, 1]^{m}$,
where $m$ is the number of motors relating to navigation and each dimension represents the command sent to a specific motor. 
The task of our method is to utilize visual and inertial perception sensors on a mobile robot for terrain classification. Our model will output a terrain category $t\in T = \{ t | t = 0, ..., n - 1\}$, where $n$ is the number of terrain types. In our setting, we only focus on surfaces that are navigable with different textures and physical properties and try to reduce the trajectory-following errors during navigation. Avoiding non-traversable surfaces like water or large rocks is not the point of the paper.

let $T$ be the set of known terrains, and the set of unknown terrains $T_u$ be the complement of $T$. Given several pre-defined trajectories on surface $i$, we define $MSE_i(p)$ as the mean squared error of the controller $p$ trying to follow those trajectories on surface $i$. From the set of known surfaces $T$, we can obtain a set of trained controller $P = \{p_i | i = 0, ..., k - 1\}$, and we have
$$\arg \min_x MSE_i(p_x) = i,$$
where $p_i$ is the controller pre-trained on the known surface $i\in T$. The ground truth category $c_i$ of surface $i$ is defined as follows:
\begin{equation}\label{eqn:classes}
\resizebox{.6 \hsize }{!}{$
c_i = 
\begin{cases}
i& if\  i \in T\\
\underset{x}{\arg \min}\ MSE_i(p_x) & if\  i \in T_u\\ 
\end{cases}
$} 
\end{equation}

In our setting, known terrains are surfaces on which our model has training data, so the perception model and nonlinear model predictive controller (NMPC) can be trained or fine-tuned on those surfaces; unknown terrains are ones that our model has never encountered during training. The ground truth categories of the terrain are defined by the characteristics of the surfaces during navigation, or specifically, by the type of controllers that has the best performance on such terrain. For known terrain, we collected image, IMU, state of the tracked robot, planned trajectories, and actual trajectories to train the perception model and controller; for unknown terrain, we collected perception data as well as the performance of each controller pre-trained on the known surfaces according to definition~\ref{eqn:classes}, for the purpose of evaluation.



\subsection{Image and IMU Backbone}

The proposed \ours{} consists of one image stream and one IMU stream, as shown in Figure~\ref{fig:network}. For the image stream, we adopted EfficientNet-B0~\cite{efficientNet} as our image feature extractor. Given an image patch $I_{patch}$, the image branch takes a series of convolution operations with kernel size of 3 and 5, and eventually down-sampled to 1280 channels with spatial resolution of $7\times7$. After flattening and two linear layers with batch normalization and drop-out, we can obtain an intermediate feature $F_{img}\in\mathbb{R}^{d}$, where $d$ is the shared dimension between $F_{img}$ and IMU feature $F_{imu}$.

Considering the effect of actions on the IMU pattern, our IMU stream takes an IMU sequence $I_{imu}$ and the corresponding PWM signal $S$ for some time duration $D$. Given the raw input sequence $H = [I_{imu}; S] \in \mathbb{R}^{D\times (6 + m)}$, the output feature $F_{attn}$ is computed as:
\begin{equation}
     F_{attn} = softmax(k(H)^T \cdot q(H))^T \cdot v(H),
     \label{eq: attention}
\end{equation}
where $k(H),\ q(H),\ v(H)$ represent the key, query, and value feature, respectively, as in the self-attention~\cite{attentionisallyouneed} literature. Before the features are passed into the classification head, we process the feature $F_{attn}$ with the IMU denoising module.

\subsection{IMU Denoising}

Due to the noisy nature of IMU data, we introduce an attention-based denoising method to sample useful features in the time dimension. The attention map $A_{map}\in\mathbb{R}^{D\times D}$ is defined as $softmax(k(H)^T \cdot q(H))$, and we can calculate the denoised IMU feature $F_{d}$ as:
\begin{equation}
     F_{d} = F_{attn}[\ \underset{i}{argsort}(-A_{map}[i][i])[0:f_s\times D]  \ ],
     \label{eq: sample}
\end{equation}
where $f_s$ is the sampling rate, and $argsort$ sort the value in increasing order and returns the corresponding indices. This step removes the features at some specific timestamp with a low attention value for denoising purposes. After passing through a few linear layers with batch normalization and flattening, we can obtain an IMU feature $F_{imu}\in\mathbb{R}^{d}$.

\subsection{Classification Head in Shared Embedding Space}

After going through the image and IMU stream, we can obtain $F_{img}$ and $F_{imu}$ with dimension $d$. We used a learnable weight for each channel of the features to fuse $F_{img}$ and $F_{imu}$. After several fully connected layers and activation, the network will output the vector with a classification score.

In the classification head, we want to train the network to learn a shared embedding space for $F_{img}$ and $F_{imu}$, which is achieved by adding a similarity constraint in the network during training. The purpose of creating a shared embedding space is to alleviate the effect of one feature taking over the other. In addition, we want to add some robustness to the network: if the data from one branch is corrupted and produced a terrible feature, it would not significantly affect and corrupt the classification head because the other branch could have some effect of averaging the corrupted features back to the shared embedding space.

\subsection{Loss Function}

\noindent\textbf{Mean-Squared-Error Loss:} To create a shared embedding space for $F_{img}$ and $F_{imu}$, we penalize the network for extracting significantly different feature vectors from different modalities of the same data point. The MSE loss is defined as follows:
\begin{equation}
     \mathcal{L}_\textrm{MSE} = \sum
      (F_{img} - F_{imu})^2,
     \label{eq: loss_mse}
\end{equation}

\noindent\textbf{Cross-Entropy Loss:} This is a standard loss for the classification task, defined as follows:
\begin{equation}
     \mathcal{L}_\textrm{CE} = - \sum_{t\in T
     } c_{t}\log(P_t
     ),
     \label{eq: loss_ce}
\end{equation}
where $T$ denotes the set of known terrain, $P_t$ denotes the output probability map corresponding to terrain $t$, and $c_t$ corresponds to the ground-truth label. We use this loss on each branch of the network as well as the final classification head.

\subsection{Adaptive \rev{Scheduling} Control Framework}

In this subsection, we give more details of our controller and present our adaptive \rev{scheduling} control framework with \ours{} on known and unknown terrain. 

\noindent\textbf{Nonlinear Model Predictive Controller:} The overall goal of the controller is to closely follow the planned trajectory, which is significantly affected by the terrain type. For known terrains, we collected some navigation data and use those to fit a nonlinear model predictive controller (NMPC) and a Gaussian Process (GP) model to learn the interaction between the tracks and the terrain.
The NMPC trajectory tracking control problem at time instance $k$ can be formulated as follows:
\begin{equation}\label{eqn:cost_plan}
    \begin{aligned}
    \min\limits_{u}~&{J(x(k),U)} = \\
    &\sum_{i=0}^{N-1}\left( {||Q x(i|k)||_{2} + ||R~u(i|k)||}_{2} + {||R_d~ \Delta u(i|k)||_2}\right)\\
    s.t.~& x(i+1|k) = x(i|k) + f\left(x(i|k),u(i|k)\right)T_{s},\\
    & x(0|k) = x_0,~x(i|k) \in \mathbf{X},\\
    & u(0|k) = u_0,~u(i|k) \in \mathbf{U},\\
    & U = col\left\{u(0|k), \dots, u(N-1|k) \right\},
    \end{aligned}
\end{equation}
where $N$ is the prediction horizon. $x(i|k)$ and $u(i|k)$ denote the predicted values of the model state and input, respectively, at time $k + i$ based on the information that is available at time $k$. $\Delta u(i|k) = u(i|k)-u(i-1|k)$ is included in the cost function to smooth the control output. The state vector is $[x_{c}, y_{c}, \theta]^T$, where $x_{c}$ and $y_{c}$ are the 2D Cartesian coordinates of the robot, and $\theta$ is the heading. The control output vector $u = [v,\omega]^T$ is composed of the linear and angular speed commands of the robot. The nonlinear prediction model $f(x, u)$ can be described by the following equations: 
\begin{equation}
    \left\{
    \begin{aligned}
     & \dot{x} = v \cdot cos\left(\theta \right)\\
     & \dot{y} = v \cdot sin\left(\theta \right).\\
     & \dot{\theta} = \omega\\
    \end{aligned}
    \right.
\label{eqn:ss_con}
\end{equation}

The output of the NMPC, $u$, is then mapped to the PWM signals $S$ controlling the two motors driving each track of the robot. The terrain track interaction\rev{, caused by terrain resistance and deformation,} is captured by the GP model: $S = gp\left(u\right)$. \rev{The GP models provide the required $S$ for each terrain, so that the actual linear and angular speeds match with the desired $u$ calculated by the NMPC.} Since $S$ is two-dimensional, corresponding to the left and right track motors, two one-dimensional GP models are fitted for each known terrain. The Radial Basis Function (RBF) kernel is used in the GP models:
\begin{equation}
    \kappa(u_i, u_j) = \sigma^2_f exp\left(- \frac{1}{2}\left(u_i-u_j\right)^TL^{-2}\left(u_i-u_j\right)\right)+\sigma^2_n
\end{equation}
where $L$ is the diagonal length scale matrix and $\sigma_f$, $\sigma_n$ represent the data and prior noise variance, and $u_i$, $u_j$ represent data features.
\begin{figure}[t]
    \centering
         \includegraphics[clip, trim=5.5cm 5.1cm 5.9cm 3.5cm, width=\linewidth]{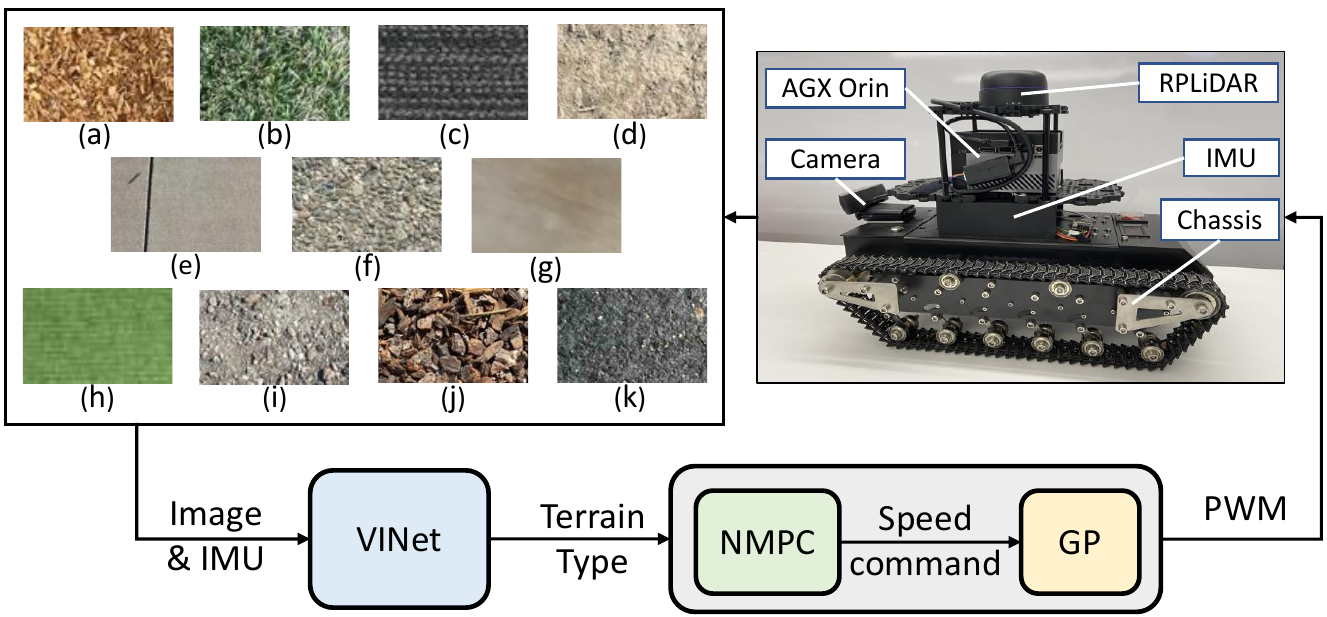}
     \caption{\textbf{Adaptive \rev{scheduling} control framework:} 
     The known terrain types (a to e) are mulch, grass, black carpet, soil, and concrete; the unknown terrain types (f to k) are small rocks, smooth tiles, green carpet, rocky surface, wood rocks, and asphalt.}
     \label{fig:robot_bunker}
     \vspace{-6mm}
\end{figure}

\noindent\textbf{Adaptive \rev{Scheduling} Control:} As shown in Figure~\ref{fig:robot_bunker}, we deploy \ours{} on a tracked robot, and our model can use visual and inertial feedback during navigation to predict a terrain class, which corresponds to the pre-trained controller and would have the best performance on this terrain if the prediction is correct. Once the robot receives the prediction, it will adjust to the suitable controller on this terrain. 

For unknown terrain, our classifier would still choose one of the categories in the original label set since our classifier is only trained on known surfaces. Because of the introduction of IMU as inertial feedback and the terrain label assignment measured by the navigation feedback from the controller, \ours{} can make a prediction based more on generalized navigation features instead of purely semantic labels, leading to a better choice of the controller on an unknown surface.
For evaluation of perception, we find a few unknown terrains and test our pre-trained controller to assign those terrains to one of our pre-trained controllers, as defined in Equation~\ref{eqn:classes}.


\subsection{Motivation for Using IMU data and Navigation-based Labeling Scheme}


While the image-based classifier can achieve a good result on standardized classification benchmarks, it does not directly translate to good performance in navigation. First, the categories in standard data sets are usually semantic-based; on the other hand, it's less intuitive to give a fair traversability or material label that is useful for navigation, and most existing practices are usually heuristic-based~\cite{offseg, geo1} for benchmarking purposes. 

Second, a purely vision-based method could easily overfit the known data and makes it harder to generalize to more unknown terrains. For example, grass and artificial grass might be visually similar, but they might have different hardness and slipperiness. Because of such difference, the controller fitted with data navigating over grass might lead to worse navigation on the artificial grass because the classifier could not distinguish their navigation features by only relying on the visual difference.

To solve those issues, we eliminate the heuristics in the labeling by using some pre-trained controllers as labels for the terrain, on which the controller has the best performance. The labeling becomes navigation-based instead of semantic-based: once our model is trained according to this labeling scheme and the model can predict correctly, the output would choose the best controller to use while navigating on the corresponding surface, which could directly benefit the navigation. 
Through navigation-based labeling, the unknown terrain will be predicted as one of the known terrains and the interpretation is that those two types of terrains could have similar navigation features due to some visual textures. The output could still be random and meaningless sometimes unless we have some feedback during the navigation. To improve the Interpretability and generalizability of the output, we add an IMU sequence as input during navigation, in addition to the visual input.



By introducing the IMU, we want to improve the model's ability to generalize on unseen terrains. The model can build connections between the extracted features from inertial sensing of IMU and the navigation features of the terrain, so the network can have less bias due to over-fitting on the images. Because all of the terrain labels are based on navigation data from known terrains, the prediction output is optimized with respect to the IMU and Image input. In Section~\ref{exp}, we validate our statement by showing comparisons with and without IMU as input to the network.


%% file: sections/result.tex
\section{Implementation Details and Analysis}
\label{exp}

\subsection{Data Collection and Controller Pre-training}
\label{data_collect}
We collected 4923 images of size $1280 \times 720$ among 11 different terrain types and the corresponding inertial sequences at 30 Hz. The 11 terrain types include black carpet, green carpet, smooth tiles, concrete, grass, soil, mulch, wood rocks, rocky surface, asphalt, and small rocks.

Gaussian Process models used in the tracking controller are fitted by the collected data on each known terrain, including the PWM signal $S$ and the corresponding robot's linear and angular speeds $u = [v, \omega]^T$ calculated based on the robot localization output. For each terrain, about 1500 data pairs $\{S, u\}$ were collected and used to train the GP model. Then, the NMPC controller is tuned manually on each terrain type to further optimize the trajectory tracking performance on each known type.
For the purpose of evaluation, we find some other surfaces as unknown terrains and evaluate those pre-trained controllers on such terrains to get the best controller as its navigation-based label. 
Due to limited space, we show the details of each terrain class and evaluated trajectories in the full report~\cite{vinet}.

\subsection{Implementation Details}

We evaluate our network with the collected data as described in~\ref{data_collect}. We use the Adam optimizer with a learning rate of 0.0003 and exponential decay of 0.95. All experiments are conducted on a single GeForce RTX 2080 Ti GPU and trained with a batch size of 128 for 200 epochs. 
We sample a patch of size $128\times 128$ along the expected trajectory based on the current velocity and heading, and extract the corresponding IMU while the robot navigates near that image patch region. For all experiments, we set the length of the IMU sequence $D$ to be 60, and the sampling rate of the IMU sequence $f_s$ to be $5/6$. We choose the dimension of the shared embedding space to be 512 and 256 in our experiments.
For evaluation, we use the averaged accuracy, which is calculated as:
\begin{equation}
     Acc = \frac{\sum_{i} 1(T(I_{img},I_{imu}) = c)}{\sum_{i} 1}
     \label{metric:acc}
\end{equation}

\subsection{Comparisons and Ablation Study}

In this section, we show some quantitative comparisons under different settings. 
For comparison methods, we used a recent method~\cite{svm_method}, which is based on SVM with principle component analysis. We implement this method according to the paper since code is not available.



\input{tables/cls5_comp}

\noindent\textbf{5 known surfaces and 6 unknown surfaces:} In Table~\ref{tab:cls5_comp}, we show the performance of several models with different parameters. The five known terrains are black carpet, concrete, grass, soil, and mulch. The IMU-only and IMG-only models are directly separated from each branch of the proposed \ours{}. We can see that utilizing both branches would have better performance than using either one of the branches. For the image backbone, using models pre-trained on ImageNet~\cite{russakovsky2015imagenet} can also boost the performance, but our method still has the best performance with or without pre-training. In addition, we discover that adding the PWM signal command to the original 6-axis IMU data can greatly improve the classification accuracy for the IMU branch. \rev{Please refer to Equ.~\ref{eqn:classes} for the definition of ground truth labels of unknown surfaces.}

\noindent\textbf{11 known surfaces and no unknown surfaces:} To illustrate the performance of \ours{} on traditional terrain classifications task, we consider all 11 surface types as known and split the types evenly to training and testing according to 7:3 ratio. As shown in Table~\ref{tab:cls11_comp}, the performance of \ours{} goes up to $98\%$ with pre-training.

\noindent\textbf{Ablation study:} In Table~\ref{tab:cls5_ablation}, we consider the effect of each component of our proposed network. We can see that removing any one of the components would lead to a performance drop, especially when the shared dimension is 256. In addition, we discover that for most of the models, except for the one without any inter. CE loss and shared embedding, using 256 for the shared embedding dimensions leads to overall better performance, especially with all components, even though the number of parameters is less.
\vspace{-2mm}

\input{tables/cls11_comp}

\input{tables/cls5_ablation}

\subsection{Navigation With Adaptive Control}



We use a tracked robot to evaluate the performance of the method, as shown in Figure~\ref{fig:robot_bunker}. The external dimension of the robot is $390\times 270\times 145 \ mm$, and the track width is $40 \ mm$.
The tracked robot has a 12V DC motor with 5 kg-cm rated torque specs on each side of the robot and it can carry a maximum payload of 30kg. 
We equipped the robot with a NVIDIA Jetson AGX Orin for processing, a single-line LiDAR (RPLiDAR-A3) for navigation, a 1080p camera (Logitech C920) for terrain image capturing, and an IMU (3DM-GX5-25) for motion data recording. All computations including data processing, inference and control are computed on the edge.

\begin{figure}[t]
\centering
\includegraphics[clip, trim=0.1cm 0.2cm 0.17cm 0.6cm, width = \linewidth]{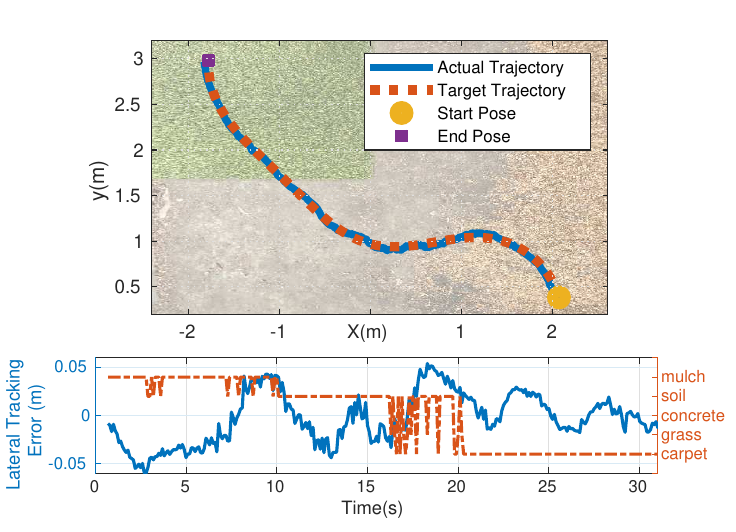}
\caption{\textbf{Trajectory tracking control result of the VINet-based adaptive controller:} Note that our classifier will classify artificial grass as carpet instead of grass, which is based on navigation properties and inertial sensing, instead of purely visual features. We tested the grass controller on artificial grass and could not record a meaningful trajectory because the tracking error is too large to reach the goal location. \rev{We see chatting classifications during transition between surfaces due to the lag between visual and IMU input.}}
\label{fig:traj_plot}
\vspace{-3mm}
\end{figure}

As shown in Figure~\ref{fig:robot_bunker}, the adaptive tracking controller receives the terrain type from the VINet and switches to the corresponding controller in real-time. We conducted multiple test cases with the adaptive controller with the same trajectories covering three different terrains, including mulch, soil as known surfaces, and artificial grass as unknown surfaces, as shown in Figure~\ref{fig:traj_plot}. For comparison, the baseline controller does not receive the terrain type output from VINet and thus uses only one controller on all three types of terrains. In the test, we choose the carpet and mulch controllers as the baseline controllers. Figure~\ref{fig:traj_plot} shows a sample trajectory tracking test case with the adaptive controller.

The tracking error of the adaptive and the baseline controllers are shown in Table~\ref{tab:track_err}. Our proposed VINet-based adaptive control framework has the best performance in terms of the root-mean-square error (RMSE) between the actual and target trajectories. 

\begin{table}[t]
\begin{center}
\begin{tabular}{cccc}
\hline
 & Adaptive & Carpet baseline & Mulch baseline\\
\hline

\begin{tabular}{@{}c@{}}Lateral RMSE (m)\end{tabular} & 0.026 & 0.050 & 0.029\\

\hline
\end{tabular}
\end{center}
\vspace{-3mm}
\caption{\textbf{Tracking Error Comparison}}
\label{tab:track_err}
\vspace{-7mm}
\end{table}

%% file: tables/cls5_comp.tex
\begin{table}[t]
\centering
\resizebox{\columnwidth}{!}{%
\begin{tabular}{cccccc}
  \toprule
  \textbf{Modality} & \textbf{IMG Backbone} & \textbf{Pre-training} & \textbf{IMU Dims} & \textbf{Shared Dims} & \textbf{Accuracy} $\uparrow$ \\
  \midrule
SVM-based~\cite{svm_method} & -  & - & - & - & 0.0282 \\ 
 \midrule
  \multirow{1}[3]{*}{IMU Branch} & -  & - & 6 & 512 & 0.4227 \\ 
        & -  & - & 6 & 256 & 0.4629 \\ 
  \midrule
  \multirow{1}[3]{*}{IMU Branch} & -  & - & 8 & 512 & 0.5072 \\ 
        & -  & - & 8 & 256 & 0.5175 \\ 
  \midrule
   \multirow{3}[4]{*}{IMG Branch}  & EfficientNet-b0 & \ding{55} & - & 512 & 0.4227 \\ 
        & EfficientNet-b0  & \ding{55} & - & 256 & 0.4198\\ 
       & EfficientNet-b0   & \checkmark & - & 512 & 0.6297\\ 
     & EfficientNet-b0     & \checkmark & - & 256 & 0.6797\\ 
  \midrule
   \multirow{3}[4]{*}{IMG + IMU (\ours{})} & EfficientNet-b0 & \ding{55}  & 6 & 512 & 0.5486 \\ 
        & EfficientNet-b0   & \ding{55}& 6 & 256 & 0.5595 \\ 
        & EfficientNet-b0  & \checkmark & 6 & 512 & 0.6642\\ 
        & EfficientNet-b0  & \checkmark & 6 & 256 & 0.6809\\ 
  \midrule
   \multirow{3}[4]{*}{IMG + IMU (\ours{})} & EfficientNet-b0  & \ding{55} & 8 & 512 & 0.5532 \\ 
         & EfficientNet-b0 & \ding{55} & 8 & 256 & 0.7171\\ 
        & EfficientNet-b0   & \checkmark & 8 & 512 & 0.6786\\ 
        & EfficientNet-b0  & \checkmark & 8 & 256 & 0.7648\\ 
  \bottomrule
\end{tabular}
}
\vspace{-1mm}

\caption{\textbf{Model comparisons under 5 known terrains setting:} All models are trained on 5 known terrains and tested on 6 unknown terrains. The image backbone are pre-trained on ImageNet~\cite{russakovsky2015imagenet}. We use 6-axis of the IMU when IMU dimension is 6, \rev{and additionally add the PWM signal command when IMU dimension is 8.}
}

\label{tab:cls5_comp}

\vspace{-7mm}
\end{table}

%% file: tables/cls11_comp.tex
\begin{table}[t]
\centering
\resizebox{\columnwidth}{!}{%
\begin{tabular}{ccccc}
  \toprule
  \textbf{Modality}  & \textbf{IMG Backbone} & \textbf{Pre-training} & \textbf{Shared Dims} & \textbf{Accuracy} $\uparrow$ \\
  
  \midrule
  SVM-based~\cite{svm_method} & -  & - & - & 0.5943 \\ 
 \midrule
  \multirow{1}[3]{*}{IMU Branch} & -  & -  & 512 & 0.7109 \\ 
        & -  & -  & 256 & 0.729 \\ 
  \midrule
   \multirow{3}[4]{*}{IMG Branch}  & EfficientNet-b0 & \ding{55}  & 512 & 0.5481 \\ 
        & EfficientNet-b0  & \ding{55}  & 256 & 0.6143\\ 
       & EfficientNet-b0   & \checkmark  & 512 & 0.9026\\ 
        & EfficientNet-b0  & \checkmark  & 256 & 0.8979\\ 
  \midrule
   \multirow{3}[4]{*}{IMG + IMU (\ours{})}  & EfficientNet-b0 & \ding{55}  & 512 & 0.779 \\ 
       & EfficientNet-b0   & \ding{55}  & 256 & 0.8348 \\ 
       & EfficientNet-b0    & \checkmark & 512 & 0.9692\\ 
       & EfficientNet-b0  & \checkmark   & 256 & 0.9837\\ 
  \bottomrule
\end{tabular}
}
\caption{\textbf{Model comparisons under 11 known terrains setting:} All models are trained and tested on 11 terrains. We only use 6-axis of the IMU when IMU dimension is 6.
}

\label{tab:cls11_comp}

\vspace{-3mm}
\end{table}

%% file: tables/cls5_ablation.tex
\begin{table}[t]
\centering
\resizebox{\columnwidth}{!}{%
\begin{tabular}{cccccc}
  \toprule
  \textbf{\shortstack{Inter. \\ CE Loss}} & \textbf{\shortstack{Shared \\Embedding}}    & \textbf{\shortstack{IMU \\Denoising}} & \textbf{\shortstack{Learned \\ Weighting}} & \multirow{-1}[2]{*}{\textbf{Shared Dims}} & \multirow{-1}[2]{*}{\textbf{Accuracy} $\uparrow$} \\
  \midrule
     \ding{55} & \ding{55}  & \checkmark  & \checkmark & 512 & 0.6803\\ 
     \ding{55} & \ding{55}  & \checkmark  & \checkmark & 256 & 0.6711\\ 
     \checkmark & \ding{55}  & \checkmark  & \checkmark  & 512 & 0.6578\\ 
     \checkmark & \ding{55}  & \checkmark  & \checkmark  & 256 & 0.6728\\ 
  \midrule
     \checkmark & \checkmark    & \ding{55} & \checkmark & 512 & 0.6515\\   
     \checkmark & \checkmark  & \ding{55}   & \checkmark  & 256 & 0.6561\\   
 \midrule
     \checkmark & \checkmark  & \checkmark  & \ding{55}  & 512 & 0.6619\\   
     \checkmark & \checkmark  & \checkmark & \ding{55}  & 256 & 0.6912\\   
 \midrule
     \checkmark & \checkmark  & \checkmark  & \checkmark  & 512 & 0.6786\\   
     \checkmark & \checkmark  & \checkmark  & \checkmark  & 256 & 0.7648\\   
  \bottomrule
\end{tabular}
}
\caption{\textbf{Ablation study on 5 known terrains setting:} The inter. CE loss is the cross-entropy loss used on the intermediate output of each branch. For shared embedding, we remove the MSE loss in the network for ablation. For ablation of IMU denosing and learned weighting, we directly use the original attention module and concatenation of the features from image and IMU. 
}

\label{tab:cls5_ablation}

\vspace{-5mm}
\end{table}







%% file: sections/conclusion.tex
\section{Conclusions, Limitations, and Future Works}

In this paper, we present a novel terrain classification method \ours{} for terrain classification and adaptive navigation control. We propose a novel IMU denoising module and a shared latent embedding space for image and IMU sequence. Our proposed navigation-based labeling and generalization can enable safe and stable navigation on unknown terrain, which is more intuitive and beneficial for navigation tasks compared to traditional semantic labels.

There are some limitations. For example, we did not extensively test our adaptive framework on more unknown terrains in a single run, since it's difficult to find multiple terrains on a single site. In the future, we want to fully exploit the controller design to better utilize the navigation-based classification scheme, and also combine it with other functionalities like traversability analysis and obstacle avoidance.


%% file: appendix/terrain_classes.tex
\section{More details on terrain classes}

In Table~\ref{tab:unknown_classes}, we give more details on the assignment of navigation-based labels for unknown classes based on known classes. On each type of unknown terrain, we sample $5-6$ trajectories for each different controller as measurement for label assignment. In particular, we give an example of assignment of rocky surface in Table~\ref{tab:rocky}. We also show one sampled trajectory for each controller on  rocky surface in Figure~\ref{fig:rocky}.

\input{tables/tab_cls}

\input{tables/rocky_surf_rmse}

%% file: tables/tab_cls.tex
\begin{table}[h]
\vspace{-5pt}
\begin{center}
\begin{tabular}{|c|c|c|}
\hline
\begin{tabular}[c]{@{}c@{}}Original Semantic-based \\ Classes\end{tabular} & \multicolumn{1}{l|}{\begin{tabular}[c]{@{}l@{}}Known / \\ Unknown\end{tabular}} & \begin{tabular}[c]{@{}c@{}}Navigation-based \\ Classes\end{tabular} \\ \hline
Black Carpet                                                               & Known                                                                           & $p_0$ - Carpet                                                              \\ \hline
Concrete                                                                   & Known                                                                           & $p_1$ - Concrete                                                            \\ \hline
Grass                                                                      & Known                                                                           & $p_2$ - Grass                                                               \\ \hline
Soil                                                                       & Known                                                                           & $p_3$ - Soil                                                                \\ \hline
Mulch                                                                      & Known                                                                           & $p_4$ - Mulch                                                               \\ \hline
Green Carpet                                                               & Unknown                                                                         & $p_0$ - Carpet                                                              \\ \hline
Rocky Surface                                                              & Unknown                                                                         & $p_1$ - Concrete                                                            \\ \hline
Wood Rocks                                                                 & Unknown                                                                         & $p_4$ - Mulch                                                               \\ \hline
Asphalt                                                                    & Unknown                                                                         & $p_3$ - Soil                                                                \\ \hline
Small Rocks                                                                & Unknown                                                                         & $p_3$ - Soil                                                                \\ \hline
Smooth Tiles                                                               & Unknown                                                                         & $p_1$ - Concrete                                                            \\ \hline
Artificial Grass*                                                           & Unknown                                                                         & $p_0$ - Carpet                                                              \\ \hline
\end{tabular}
\end{center}

\caption{\textbf{Terrain labels for known and unknown terrains:} The known classes are labeled as their original classes and the corresponding controller, while the unknown terrains are labeled based on the performance of each controller trained on each known terrain. * indicates terrain only tested in navigation, but not included in the perception benchmark.}
\label{tab:unknown_classes}

\end{table}

%% file: tables/rocky_surf_rmse.tex
\begin{table}[t]
\centering
\resizebox{0.86\columnwidth}{!}{%
\begin{tabular}{c|ccc}
  \toprule
Controller & Overall  & Longtitude  & Latitude \\ 
Type & RMSE  & RMSE  & RMSE \\ 
  \midrule
$p_0$ - Carpet       &     0.045516  &   0.034005  &  0.030255   \\
\textbf{$p_1$ - Concrete}       &     \textbf{0.032945}  &  \textbf{0.031562}  &  \textbf{0.009446}   \\
$p_2$ - Grass       &     0.175690   & 0.077231  &    0.157800   \\
$p_3$ - Soil       &     0.045092   & 0.037987   & 0.024296   \\
$p_4$ - Mulch       &     0.066245  &  0.061325   & 0.025052   \\
\bottomrule
\end{tabular}
}
\caption{\textbf{Tracking errors on rocky surface:} We sample $5-6$ trajectories on this terrain (rocky surface) and measure the tracking error of each controller. We assign a controller label with least root mean squared error (RMSE) to this terrain as label, which is $p_1$ - concrete in this case.}
\label{tab:rocky}
\end{table}